\pdfoutput=1

\documentclass[11pt]{article}

\usepackage{emnlp2021}

\usepackage{times}
\usepackage{latexsym}

\usepackage[T1]{fontenc}

\usepackage[utf8]{inputenc}

\usepackage{microtype}

\usepackage{CJKutf8} 
\usepackage{amsmath}
\usepackage{amssymb}
\usepackage{bm}
\usepackage{graphicx}
\usepackage{subcaption}
\usepackage{booktabs}
\usepackage{multirow}
\usepackage{makecell}
\usepackage{enumitem}
\usepackage{todonotes}
\newcommand{\mbf}[1]{\mathbf{#1}}

\title{Improving Zero-Shot Cross-Lingual Transfer \\ Learning via Robust Training}

\author{Kuan-Hao Huang, Wasi Uddin Ahmad, Nanyun Peng, Kai-Wei Chang \\
        University of California, Los Angeles \\ 
        \texttt{\{khhuang, wasiahmad, violetpeng, kwchang\}@cs.ucla.edu}}

\begin{document}
\maketitle

\begin{abstract}

Pre-trained multilingual language encoders, such as multilingual BERT and XLM-R, show great potential for zero-shot cross-lingual transfer. 
However, these multilingual encoders do not precisely align words and phrases across languages.
Especially, learning alignments in the multilingual embedding space usually requires sentence-level or word-level parallel corpora, which are expensive to be obtained for low-resource languages. 
An alternative is to make the multilingual encoders more robust; when fine-tuning the encoder using downstream task, we train the encoder to tolerate noise in the contextual embedding spaces such that even if the representations of different languages are not aligned well, the model can still achieve good performance on zero-shot cross-lingual transfer.
In this work, we propose a learning strategy for training robust models by drawing connections between adversarial examples and the failure cases of zero-shot cross-lingual transfer.
We adopt two widely used robust training methods, adversarial training and randomized smoothing, to train the desired robust model. The experimental results demonstrate that robust training improves zero-shot cross-lingual transfer on text classification tasks. The improvement is more significant in the \emph{generalized} cross-lingual transfer setting, where the pair of input sentences belong to two different languages.

\end{abstract}

\section{Introduction}

Zero-shot cross-lingual transfer learning aims to learn models with data available in one or more source languages and use them in other target languages for which there is no data (zero-resource) available. The zero-shot cross-lingual transfer has a great practical value for low-resource languages since it reduces the requirement of labeled data to learn models for downstream tasks, e.g., text classification \cite{Conneau18xnli,Yang19pawsx} and question answering \cite{Lewis20mlqa}.

Recently, pre-trained multilingual language encoders, such as multilingual BERT \cite{Devlin19bert} and XLM-R \cite{Conneau20xlmr}, demonstrate promising performance on zero-shot cross-lingual transfer learning for a wide range of downstream tasks \cite{Hu20xtreme,Liang20xglue}.
These language encoders learn a shared multilingual contextual embedding space; they are able to represent word pairs in parallel sentences with similar contextual representations.
However, the multilingual encoders fail to capture this similarity when the source and target languages are less similar at levels of morphology, syntax, and semantics \cite{ahmad2019cross,Ahmad19difficult}.

\begin{figure*}[!t]
\centering
\begin{subfigure}[b]{0.45\textwidth}
    \centering
    \includegraphics[width=\textwidth]{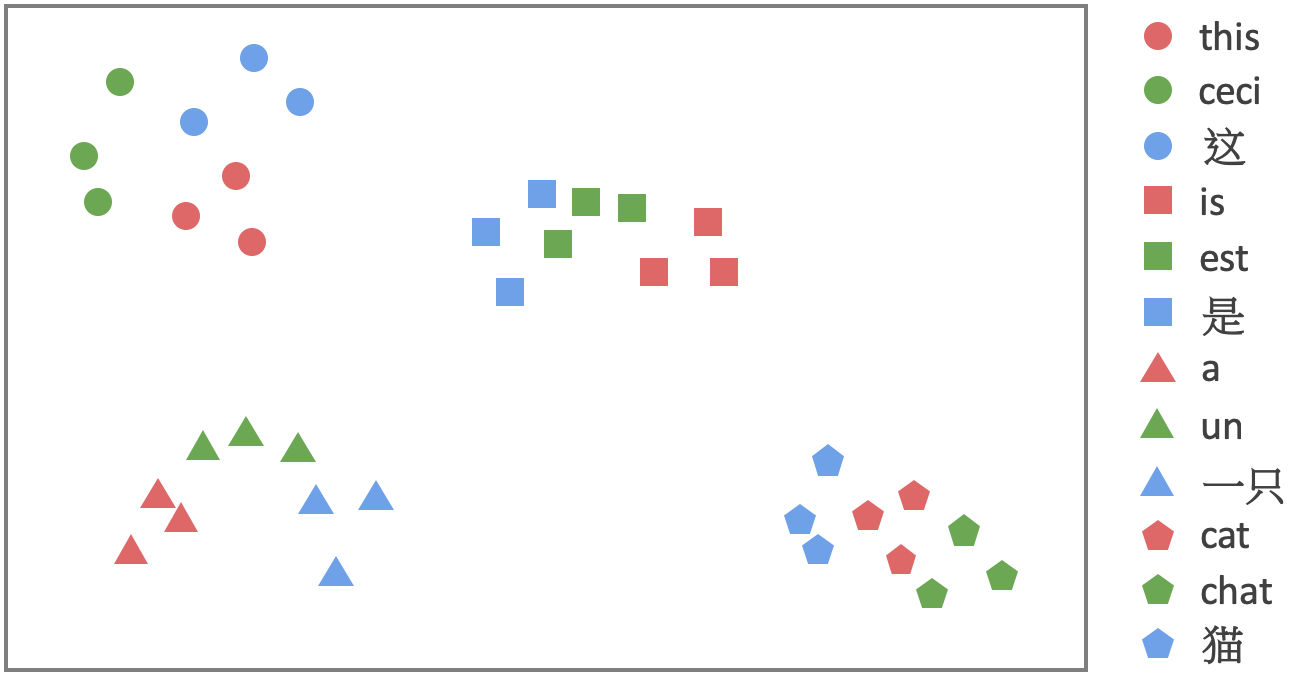}
    \caption{Contextual representations of different words.}
    \label{fig:embed1}
\end{subfigure}
\hspace{1.5em}
\begin{subfigure}[b]{0.45\textwidth}
    \centering
    \includegraphics[width=\textwidth]{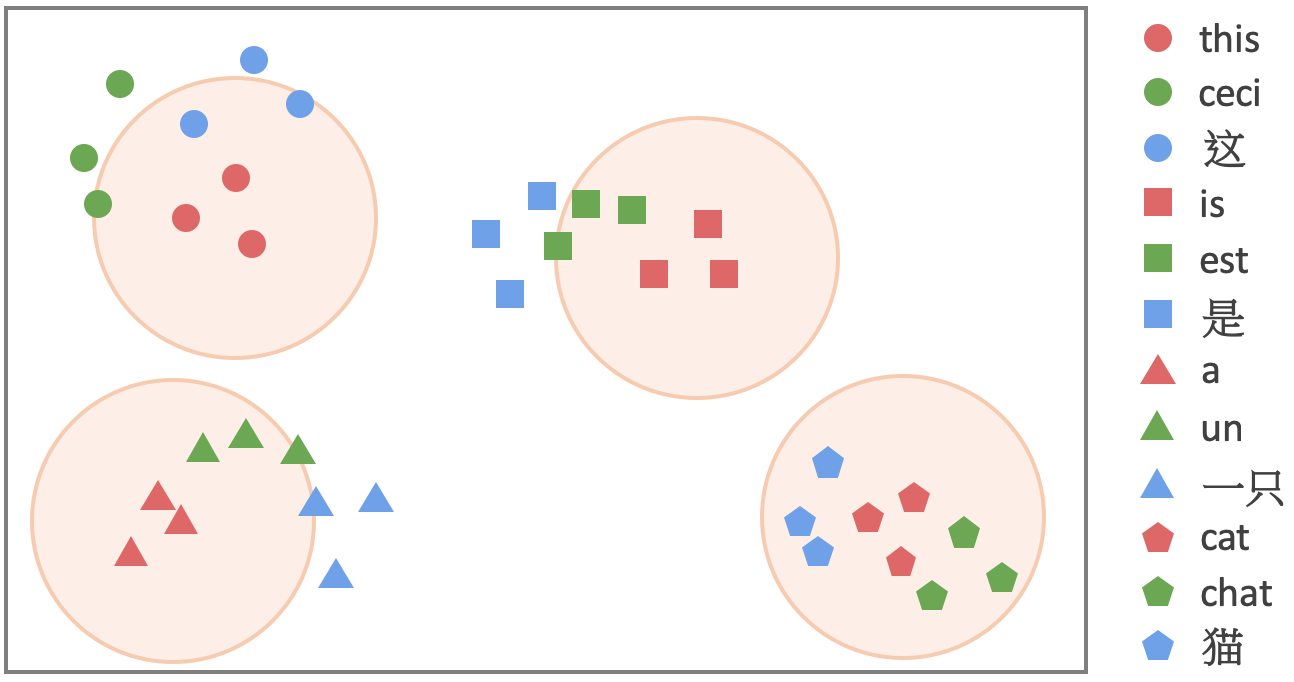}
    \caption{Robust regions try to cover neighbor embeddings.}
    \label{fig:embed2}
\end{subfigure}

\caption{An illustration of different words in the multilingual contextual embedding space. (a) Words with similar meanings in different languages have similar representations but they are not exactly aligned. (b) We aim to learn a robust classifier whose robust regions (orange circles) that cover as many neighbor words as possible.}
\label{fig:model}
\end{figure*}

\begin{CJK*}{UTF8}{gbsn}
Prior studies \cite{Cao20align1,Pan21align2,Dou21align3} have shown that aligning the representations of different languages in the multilingual embedding space plays an important role for zero-shot cross-lingual transfer learning. As illustrated in Figure~\ref{fig:embed1}, words with similar meanings (e.g. \emph{this}, \emph{ceci}, and \emph{这}) have similar representations in the contextual multilingual embedding space, even though these words are in different languages. This alignment helps models transfer the learned knowledge from source languages to target languages. 
Therefore, several works focus on improving the quality of alignments in the multilingual embedding space \cite{Cao20align1,Chi20align4,Pan21align2,Dou21align3}. 
Nevertheless, learning such alignments usually requires sentence-level or word-level parallel corpora, which are expensive to be obtained for low-resource languages.
In addition, because the meanings of words in different languages are usually not exactly matched, learn a perfect alignment could be impossible.
\end{CJK*}

In this work, we start from another point of view to improve zero-shot cross-lingual transfer performance. We aim to make the multilingual encoders robust such
that they can tolerate a certain amount of noise in the input embeddings. 
More specifically, as shown in Figure~\ref{fig:embed2}, we target to construct \emph{robust regions} (orange circles) for embeddings in the multilingual embedding space. During training, the robust model is expected to output similar predictions for embeddings in the same robust region. 
Therefore, as long as similar words in different languages fall into the same robust region, even if they are not perfectly aligned, the model can still have similar predictions for them.

To learn the robust model, we first draw connections between adversarial examples \cite{Li20attack1,Garg20attack2,Jin20attack3} and the failure cases of zero-shot cross-lingual transfer, and then study two widely used robust training methods to learn the robust model: (1) adversarial training \cite{Goodfellow15advtrain1,Madry18advtrain2} and (2) randomized smoothing \cite{Cohen19randomized,Ye20safer}. 
Both of them can make the model robust against perturbations in the input embeddings by modifying the training objective when fine-tuning model for the downstream task. For randomized smoothing, we also adopt the data augmentation approach \cite{Ye20safer} to learn the robust model.

We perform experiments on two cross-lingual text classification tasks, paraphrase identification and natural language inference\footnote{Our code is available at \url{https://github.com/uclanlp/Robust-XLT}}.
The experimental results demonstrate that robust training indeed improves the performance of zero-shot cross-lingual transfer on the classification benchmarks: PAWS-X \cite{Yang19pawsx} and XNLI \cite{Conneau18xnli}. On average the cross-lingual transfer performance improves by 2.1 and 1.6 points on PAWS-X and XNLI, respectively.
In addition, we show that robust training remarkably improves \emph{generalized} cross-lingual transfer \cite{Lewis20mlqa}. 
In this setting, the pair of input sentences in the text classification tasks belong to two different languages, e.g., paraphrase prediction for a pair of sentences in English and Korean.

\section{Related Work}

\paragraph{Zero-shot cross-lingual transfer learning.}
In recent years, several pre-trained multilingual language models are proposed for zero-shot cross-lingual transfer, including multilingual BERT \cite{Devlin19bert}, XLM \cite{ConneauL19xlm}, and XLM-R \cite{Conneau20xlmr,Goyal21xlmrxl}. Many studies put attentions on the rationales that make zero-shot cross-lingual transfer work \cite{K20mberts1,Lauscher20mberts2,Conneau20mberts3,Artetxe20mberts4,Dufter20mberts5}. 
Various tasks and datasests are presented to facilitate zero-shot cross-lingual transfer learning \cite{Conneau18xnli,Yang19pawsx,Clark20tydi,Artetxe20mberts4,Lewis20mlqa}. XTREME \cite{Hu20xtreme} and XGLUE \cite{Liang20xglue} further provide benchmarks for zero-shot cross-lingual transfer learning.

\paragraph{Embedding space alignments.}
Learning to align embedding spaces have always been an important research topic to improve multilinguality. Early works focus on word embedding spaces \cite{Mikolov13word1,Smith17word2,Artetxe17word3}. Recently, many approaches are proposed to align contextual word embedding spaces, such as learning rotation projections \cite{Schuster19rotate1,Aldarmaki19rotate2,Conneau20mberts3} and fine-tuning pre-trained multilingual language models \cite{Chi20align4,Feng20align5,Cao20align1,Qin20codesw1,Liu20codesw2,Dou21align3,Wei21align6}. However, most of them require additional supervision signals, such as parallel sentence pairs \cite{Chi20align4,Feng20align5,Wei21align6}, bilingual dictionary \cite{Cao20align1,Qin20codesw1,Liu20codesw2}, or both \cite{Pan21align2}. These additional supervised corpora are usually expensive for low-resource languages.

\paragraph{Embedding misalignment handling.}
Instead of directly aligning the representations, there is a line of research making the model be aware of the embedding misalignment issues by considering additional syntactic features, such as part-of-speech \cite{Kozhevnikov13pos1} and dependency parse trees \cite{Ahmad19difficult,Subburathinam19clgcn,Zhang19dep1,Liu19dep2,Ahmad21syntaxbert,Ahmad21gate}, and other syntactic features~\cite{meng2019target}.
However, those syntactic features require large human efforts to obtained.

\paragraph{Robust training.}
Recently, adversarial attacks are presented to check the robustness of NLP models, such as character manipulation \cite{Ebrahimi18hotflip,Gil19hotflip2}, word replacements \cite{Alzantot18synonym,Li20attack1,Garg20attack2,Jin20attack3}, and syntactic rearrangements \cite{Iyyer18scpn}. To against those attacks, various robust training methods are proposed. For example, \citet{Alzantot18synonym} trains a robust model by data augmentation with generated adversarial examples. Other works \cite{Ebrahimi18hotflip,Dong20convex,Zhou21dir} consider adversarial training, which includes the adversarial accuracy to the training objective. A few studies propose transformations on inputs before feeding them to models \cite{Edizel19roboenc2,Jones20roboenc}. Randomized smoothing \cite{Cohen19randomized,Ye20safer} is presented to make models robust against noise in input representations. Another line of research aims at providing theoretical guarantee of robustness, including interval bound propagation methods \cite{Jia19certify1,Huang19certify2} and verification methods \cite{Shi20verify}. Most of those robust training methods focus on defending adversarial attacks, while we propose to apply robust training methods to improve the zero-shot cross-lingual transfer performance.
\section{Zero-Shot Cross-Lingual Transfer with Robust Training}
\label{sec:method}

In this work, we focus on zero-shot cross-lingual transfer for text classification tasks. 
Our goal is to learn a classifier $f$ from a set of training examples in source languages $X_{src} = \{ (x_i, y_i)\}_{i=1}^N$. At test time, we directly use the classifier $f$ to conduct inference on a set of test examples in target languages $X_{tgt} = \{ x_i \}_{i=1}^M$. We expect that the classifier $f$ can transfer the learned knowledge from the source languages to the target languages.

\subsection{Connection with Adversarial Examples}
The aligned representations of different languages have been shown as a crucial factor \cite{Cao20align1,Chi20align4,Pan21align2} for multilingual embeddings to be effective for zero-shot cross-lingual transfer. For example, assuming the source language and the target language are English and French, respectively, and considering a pair of parallel sentences ``\textit{this is a cat}'' (in English) and ``\textit{Ceci est un chat}'' (in French), we can get the contextual representations of the source English sentence $\mbf E_{src} = (\mbf v_1, \mbf v_2, \mbf v_3, \mbf v_4)$ and the target French sentence $\mbf E_{tgt} = (\mbf u_1, \mbf u_2, \mbf u_3, \mbf u_4)$. Let $\bm{\delta}$ denote the difference between the source and the target contextual representations as follows.\footnote{For the ease of describing our idea, we assume the word orders in different languages are the same. Later in experiments, we relax this condition and present a preliminary study on the influence of word orders in Section~\ref{sec:syntax}.}
\begin{align*}
\bm{\delta} &= \mbf E_{src} - \mbf E_{tgt}  \\
&= (\mbf v_1-\mbf u_1, \mbf v_2-\mbf u_2, \mbf v_3-\mbf u_3, \mbf v_4-\mbf u_4) \\
&= (\bm{\delta}_1, \bm{\delta}_2, \bm{\delta}_3, \bm{\delta}_4). 
\end{align*}
Since words with similar meanings have similar representations, the norm of their differences $\| \bm{\delta}_i \|$ is supposed to be small. Therefore, if $f(\mbf E_{src}) = c$, we have a high probability for $f(\mbf E_{tgt}) = c$ as well, which means that the classifier $f$ is able to transfer the learned knowledge from the source language to the target language. If unfortunately, the transfer fails, we have
\begin{equation}\begin{aligned}
f(\mbf E_{tgt}) &= f(\mbf E_{src}+ \bm{\delta}) \neq f(\mbf E_{src}),  \\
\text{where} \, &\| \bm{\delta}_i \| \, \text{is small}.
\label{eq:adv}
\end{aligned}\end{equation}

We observe that Eq.~\eqref{eq:adv} is very similar to the definition of \emph{adversarial examples} \cite{Alzantot18synonym,Li20attack1,Garg20attack2,Jin20attack3}. The goal of adversarial examples is to find a small perturbation $\bm{\Delta}$ for an instance $\mbf x$ such that a classifier $h$ changes the prediction on $\mbf x$, as illustrated by the following equation.
\begin{equation}\begin{aligned}
h(\tilde{\mbf x}) &= h(\mbf x+ \bm{\Delta}) \neq h(\mbf x), \\
\text{where} &\, \| \bm{\Delta} \| \, \text{is small}.
\end{aligned}\end{equation}

For the case that cross-lingual transfer fails, the difference between the source and target representations $\bm{\delta}$ behaves like an adversarial perturbation. This inspires us to consider robust training methods, which are designed for defending adversarial examples, to improve the zero-shot cross-lingual transfer performance. More specifically, our goal is to train a robust classifier that can tolerate small perturbations on input embeddings. As shown in Figure~\ref{fig:embed2}, we aim to train a robust classifier $f$ that has robust regions (orange circles) such that the robust classifier $f$ outputs similar values for input embeddings are in the same robust region.

We study two widely used robust training methods in literature: (1) adversarial training and (2) randomized smoothing, as they have been successfully used for defending adversarial attacks \cite{Ebrahimi18hotflip,Jia19certify1,Huang19certify2,Cohen19randomized}.

\subsection{Adversarial Training}
The main idea of adversarial training is considering the most effective adversarial perturbation in each optimization iteration. More precisely, in normal training, we learn a classifier $f$ by solving the following optimization problem
\[
\min\limits_f \sum\limits_{(x,y) \in X_{src}} \mathcal{L} (f(\text{Enc}(x)), y),
\]
where $\text{Enc}(\cdot)$ is the multilingual encoder and $\mathcal{L}$ is the cross-entropy loss.
When considering adversarial training, we solve the following min-max optimization problem instead
\[
\min\limits_f \sum\limits_{(x,y) \in X_{src}} \max\limits_{\| \bm{\delta}_i \| \leq \varepsilon} \mathcal{L}(f(\text{Enc}(x)+\bm{\delta}), y),
\]
where $\varepsilon$ is a hyper-parameter to control the size of robust regions which are described by several norm balls $\| \bm{\delta}_i \|$. The inner maximization finds the most effective perturbation to change the prediction, while the outer minimization tries to ensure the correct prediction against the perturbation. With this min-max optimization, the classifier $f$ is aware of perturbations within the robust regions $\| \bm{\delta}_i \|$ and becomes more robust.

\subsection{Randomized Smoothing}

Unlike adversarial training, which always considers the most effective perturbation, randomized smoothing focuses on the expectation case and aims to guarantee the local smoothness of the classifier at the same time. Following previous work \cite{Cohen19randomized,Ye20safer}, we let $f$ be the classifier learned by solving the normal optimization problem and learn a smoothed classifier $g$ such that
\[
g(\text{Enc}(x)) = \arg \max\limits_{c \in \mathcal{Y}} \mathbb{P}_{\bm{\delta}}(f(\text{Enc}(x)+\bm{\delta})=c),
\]
where $\mathbb{P}_{\bm{\delta}}$ is a prior distribution of the perturbation $\bm{\delta}$ and $\mathcal{Y}$ is the label space.
In other words, we want that $g(\text{Enc}(x))$ has a similar output value (label predictions) to $f(\text{Enc}(x))$. The random perturbation $\bm{\delta}$ is introduced to ensure the local smoothness of $g$. That is, $g(\text{Enc}(x)+\bm{\delta})$, the output for the perturbed input, is similar to the output value of $g(\text{Enc}(x))$. Compared to the original classifier $f$, the smoothed classifier $g$ is more robust against local perturbations.

We consider two different ways to learn the smoothed classifier $g$: (1) random perturbation and (2) data augmentation.

\paragraph{Random perturbation (RP).}
Specifically, we focus on the following objective
\[
\min\limits_g \sum\limits_{(x,y) \in X_{src}} \mathbb{P}_{\bm{\delta}}( \mathcal{L}(g(\text{Enc}(x)+\bm{\delta}), y)).
\]
In each optimization step, we randomly sample a perturbation $\bm{\delta}$ from $\mathbb{P}_{\bm{\delta}}$ and add it to $\text{Enc}(x)$. Then, we use the perturbed representation as the input to calculate the loss and update the classifier $g$.

\paragraph{Data augmentation (DA).}
Another common way to approximate the smoothed classifier $g$ is data augmentation \cite{Ye20safer}. Instead of randomly sampling the perturbation $\bm{\delta}$, we consider a predefined synonym set \cite{Alzantot18synonym}. For every example $x = (w_1, w_2, ..., w_n)$ in $X_{src}$, we generate $m$ augmented examples by replacing each word $w_i$ in $x$ with one of its synonym words (including $w_i$ itself). We allow multiple replacements in one example. Then, we use the augmented data to train a smoothed classifier $g$.

It is worth noting that the predefined synonym set is required for \emph{only} source languages. Unlike previous work \cite{Qin20codesw1,Liu20codesw2}, which uses bilingual dictionary of \emph{both} source languages and target languages, the proposed method does not need any additional annotations of target languages.

\begin{table*}[t]
\centering
\setlength{\tabcolsep}{7pt}
\begin{tabular}{lcccccccc}
    \toprule
    Model             & en   & de   & es   & fr   & ja   & ko   & zh   & avg. \\
    \midrule
    mBERT*            & 94.0 & 85.7 & 87.4 & 87.0 & 73.0 & 69.6 & 77.0 & 82.0 \\
    mBERT (reproduce) & 93.7 & 85.4 & 88.2 & 87.8 & 75.3 & 74.2 & 79.1 & 83.4 \\
    mBERT-ADV         & 93.7 & \underline{86.5} & 88.5 & 87.8 & \underline{76.1} & \underline{75.3} & \underline{80.4} & \underline{84.0} \\
    mBERT-RS-RP       & \underline{\textbf{94.5}} & \underline{87.4} & \underline{\textbf{90.0}} & \underline{\textbf{89.5}} & \underline{77.9} & \underline{77.5} & \underline{\textbf{82.0}} & \underline{\textbf{85.5}} \\
    mBERT-RS-DA       & 93.5 & \underline{\textbf{87.8}} & 88.8 & \underline{88.8} & \underline{\textbf{79.3}} & \underline{\textbf{78.3}} & \underline{81.5} & \underline{85.4} \\
    \bottomrule
\end{tabular}
\caption{Averaged results of zero-shot cross-lingual transfer on PAWS-X with 10 different random seeds. Highest scores are in bold. Underlines denote that the improvement is significant with $p \leq 0.05$ for  the bootstrapped paired $t$-test. *We report the numbers in the previous paper \cite{Hu20xtreme}.}
\label{tab:pawsx}
\end{table*}

\begin{table*}[t]
\centering
\setlength{\tabcolsep}{7pt}
\begin{tabular}{l c c c c c c c c}
    \toprule
    Model     & en   & ar   & bg   & de   & el   & es   & fr   & hi \\
    \midrule
    mBERT*            & 80.8 & 64.3 & 68.0 & 70.0 & 65.3 & 73.5 & 73.4 & 58.9 \\
    mBERT (reproduce) & 82.3 & 64.8 & 68.2 & 70.8 & 66.4 & 74.3 & 73.7 & 59.7 \\
    mBERT-ADV         & 81.9 & 64.9 & 68.3 & \underline{71.7} & 66.5 & 74.4 & 74.5 & 59.6 \\
    mBERT-RS-RP       & \textbf{82.6} & \underline{65.4} & 68.7 & 70.5 & \underline{67.2} & \underline{\textbf{75.0}} & 74.1 & 59.8 \\
    mBERT-RS-DA       & 81.0 & \underline{\textbf{66.4}} & \underline{\textbf{69.9}} & \underline{\textbf{71.8}} & \underline{\textbf{68.0}} & 74.7 & \textbf{74.2} & \underline{\textbf{62.7}} \\ 
    \midrule
    Model     & ru   & sw   & th   & tr   & ur   & vi   & zh   & avg. \\
    \midrule
    mBERT*            & 67.8 & 49.7 & 54.1 & 60.9 & 57.2 & 69.3 & 67.8 & 65.4 \\
    mBERT (reproduce) & 68.7 & 50.0 & 53.0 & 60.9 & 57.7 & 70.3 & 69.2 & 66.0 \\
    mBERT-ADV         & 68.8 & 48.8 & 50.6 & 61.7 & \underline{59.2} & 70.0 & 69.4 & 66.0 \\
    mBERT-RS-RP       & \underline{69.5} & 48.4 & 50.5 & 59.7 & 57.9 & 70.5 & \underline{69.7} & 66.0 \\
    mBERT-RS-DA       & \underline{\textbf{70.6}} & \underline{\textbf{51.1}} & \underline{\textbf{55.7}} & \underline{\textbf{62.9}} & \underline{\textbf{60.9}} & \underline{\textbf{71.8}} & \underline{\textbf{71.4}} & \underline{\textbf{67.6}} \\
    \bottomrule
\end{tabular}
\caption{Averaged results of zero-shot cross-lingual transfer on XNLI with 10 different random seeds. Highest scores are in bold. Underlines denote that the improvement is significant with $p \leq 0.05$ for  the bootstrapped paired $t$-test. *We report the numbers in the previous paper \cite{Hu20xtreme}.}
\label{tab:xnli}
\end{table*}


\section{Experiments}
We conduct experiments to verify that robust training indeed improves the performance of zero-shot cross-lingual transfer. 

\subsection{Setup}

We consider two cross-lingual text classification datasets: Cross-lingual Paraphrase Adversaries from Word Scrambling (PAWS-X) \cite{Yang19pawsx} and Cross-lingual Natural Language Inference (XNLI) \cite{Conneau18xnli}. The goal of PAWS-X is to determine whether two sentences are paraphrases to each other or not. XNLI is designed for natural language inference; given a premise and a hypothesis, the classifier predicts the relation of the two sentences from \{\emph{entailment}, \emph{neutral}, \emph{contradiction}\}. 

For both datasets, we consider English as the source language and treat other languages as the target languages. We use the train, validation, and test splits provided by XTREME framework \cite{Hu20xtreme}. Specifically, we conduct 10 runs of experiments with 10 different random seeds. In each run, we train the classifier on the English training set, use the English validation set to search the best parameters, and record the results of the test sets. Finally, the averaged results of 10-run experiments are reported.

\paragraph{Compared models.} We consider the following four different models: 
\begin{itemize}[topsep=3pt, itemsep=-3pt, leftmargin=15pt]
    \item \textbf{mBERT}: the standard multilingual BERT \cite{Devlin19bert}.
    \item \textbf{mBERT-ADV}: multilingual BERT with adversarial training.
    \item \textbf{mBERT-RS-RP}: multilingual BERT with randomized smoothing via random perturbation.
    \item \textbf{mBERT-RS-DA}: multilingual BERT with randomized smoothing via data augmentation.
\end{itemize}

\begin{figure*}[ht]
\centering
\begin{subfigure}[b]{0.37\textwidth}
    \centering
    \includegraphics[width=\textwidth]{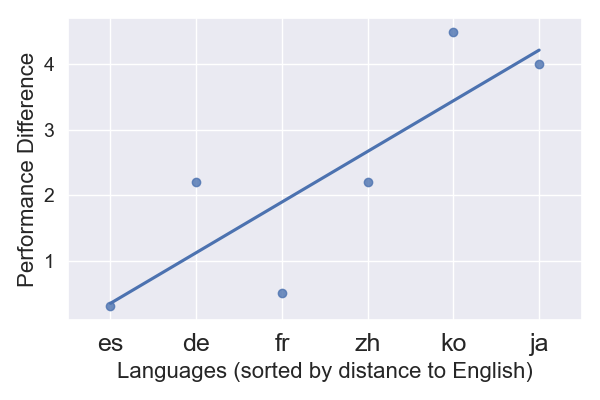}
    \caption{PAWS-X}
    \label{fig:gap_pawsx}
\end{subfigure}
\hfill
\begin{subfigure}[b]{0.615\textwidth}
    \centering
    \includegraphics[width=\textwidth]{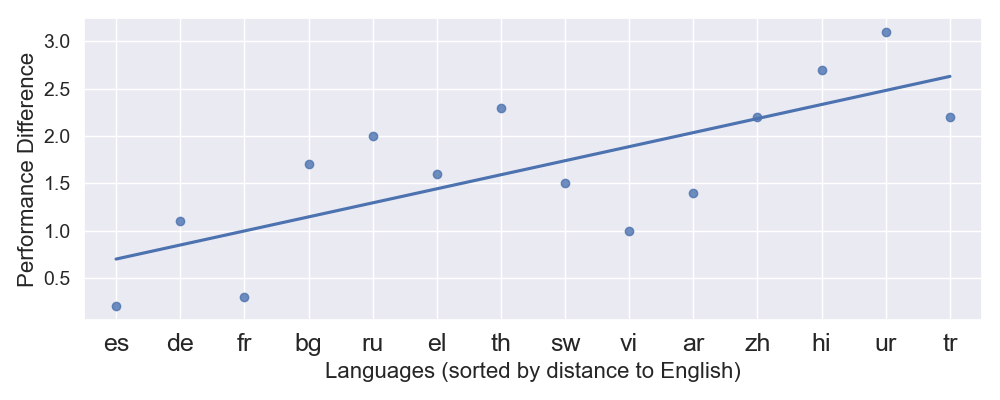}
    \caption{XNLI}
    \label{fig:gap_xnli}
\end{subfigure}
\caption{Performance difference between mBERT-RS-DA and mBERT over different languages. We sort the languages according to their distances to English from left (small) to right (large). Performance on languages with larger distances to English is improved more with the robust training.}
\label{fig:gap}
\end{figure*}

\paragraph{Implementation details.}
For adversarial training, we consider $L_{\infty}$-norm as the norm of perturbation $\| \bm{\delta}_i \|$. The size of robust regions is searched from $\{ 0.001, 0.01, 0.1, 1.0\}$. For the randomized smoothing via random perturbation, we consider uniform distribution over a $L_{\infty}$-norm ball. The size of ball is searched from $\{ 0.001, 0.01, 0.1, 1.0\}$. For the randomized smoothing via data augmentation, we consider the synonym set provide by previous work \cite{Alzantot18synonym}, which is constructed by searching nearest neighbors of words in the GloVe embedding space \cite{Pennington14glove} post-processed by the counter-fitting method \cite{Mrksic16counter}. The number of augmented examples $m$ is set to 10 and 3 for PAWS-X and XNLI, respectively, while more discussion on $m$ is shown in Section~\ref{sec:results}. For other parameters, such as the learning rate and the batch size, we follow the training scripts provided by XTREME framework \cite{Hu20xtreme}.

\subsection{Zero-Shot Cross-Lingual Transfer}
\label{sec:results}

Table~\ref{tab:pawsx} shows the averaged results of PAWS-X with 10 different random seeds. We first notice that all mBERT-ADV, mBERT-RS-RP, and mBERT-RS-DA perform better than the standard mBERT on average. Especially, robust training leads to up to 4.0\% improvement on Japanese, up to 4.1\% improvement on Korean, and up to 2.9\% improvement on Chinese. The results suggest that robust training helps in improving the performance of zero-shot cross-lingual transfer learning.

We observe that randomized smoothing is usually better than adversarial training. The reason is that adversarial training always considers the most effective adversarial perturbation during the optimization process. Adversarial perturbations are suitable for defending adversarial examples as they are specifically designed for attacking the classifier. However, in the zero-shot cross-lingual transfer case, the perturbations are not explicitly designed but reflect the natural difference between languages. Therefore, randomized smoothing, which considers the average case, becomes the better choice.

We have a similar conclusion for the XNLI dataset. As shown in Table~\ref{tab:xnli}, robust training indeed leads to improvements on zero-shot cross-lingual transfer. Again, randomized smoothing performs better than the adversarial training approach.

Finally, we compare the two different ways (random perturbation and data augmentation) to learn the smoothed classifier. They have competitive performance on PAWS-X; however, data augmentation performs better than random perturbation on XNLI. We hypothesize that the ideal robust regions in practice may not be perfect norm balls. In fact, they are more like convex hulls composed by the neighbor words \cite{Dong20convex}. By considering a predefined synonym set, mBERT-RS-DA can better capture the shapes of robust regions, leading to a more stable performance.

\paragraph{What languages are benefited most from robust training?}
We notice that cross-lingual transfer to some languages is significantly improved by robust training, especially those languages that are quite different from the source language (English). To verify this conjecture, we consider lang2vec \cite{Littell17lang2vec}, a tool that extracts features of different languages by querying the URIEL typological database\footnote{\url{http://www.cs.cmu.edu/~dmortens/projects/7\_project}}, to calculate the distance between English and other languages. Then, we show the performance gaps between mBERT-RS-DA and mBERT over all languages as well as the least square regression line in Figure~\ref{fig:gap}. Note that the languages are sorted according to their distances to English from left to right.

From Figure~\ref{fig:gap_pawsx}, we observe an obvious trend for PAWS-X that languages with larger distances to English have more performance gain with robust training. We posit that it is because languages with larger distances have more different representations from English in the multilingual embedding space. The norm of the perturbation $\bm{\delta}$ defined in Section~\ref{sec:method} will be larger and thus the failure cases occur more often. By performing robust training, we reduce failure cases that lead to a larger improvement. 
Similar trend can be observed for XNLI (Figure~\ref{fig:gap_xnli}). Performance on languages with larger distances to English is improved more with the robust training.

\begin{figure}[!t]
    \centering
    \includegraphics[width=0.98\columnwidth]{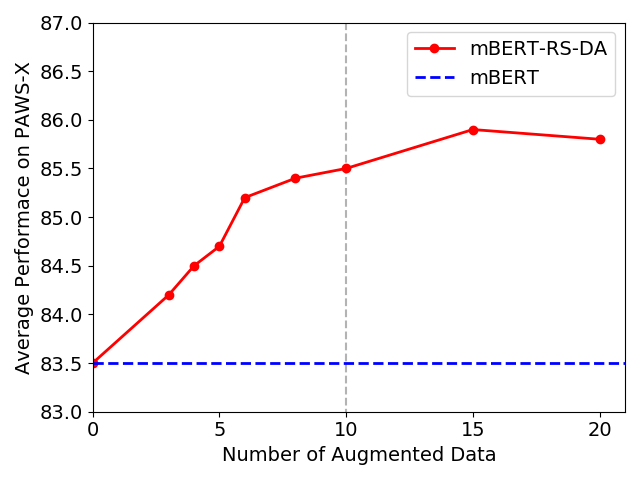}
    \caption{Performance of mBERT-RS-DA on PAWS-X over different $m$ (the number of augmented instances generated by synonym replacements). }
    \label{fig:aug}
\end{figure}

\begin{figure*}[!t]
\centering
\begin{subfigure}[b]{0.43\textwidth}
    \centering
    \includegraphics[width=\textwidth]{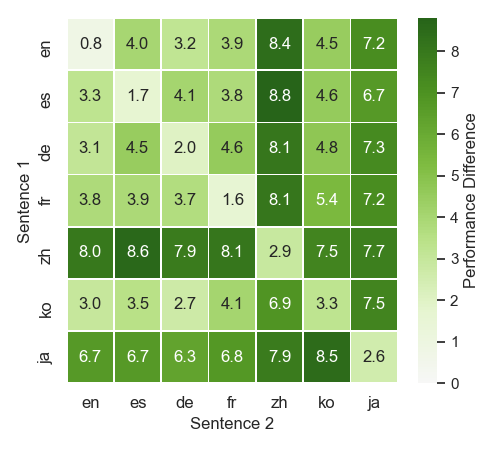}
    \caption{mBERT-RS-RP}
    \label{fig:heat-pawsx-aug}
\end{subfigure}
\hspace{2em}
\begin{subfigure}[b]{0.43\textwidth}
    \centering
    \includegraphics[width=\textwidth]{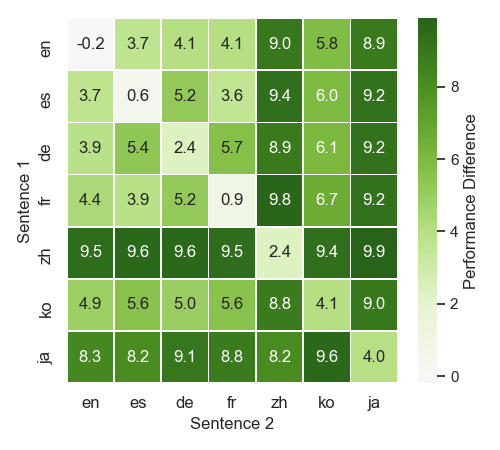}
    \caption{mBERT-RS-DA}
    \label{fig:heat-pawsx-radv}
\end{subfigure}
\caption{Results for generalized zero-shot cross-lingual transfer on PAWS-X. We report the performance difference between the compared model and mBERT over different combinations of languages.}
\label{fig:heat-pawsx}
\end{figure*}

\begin{figure*}[!t]
\centering
\begin{subfigure}[b]{0.49\textwidth}
    \centering
    \includegraphics[width=\textwidth]{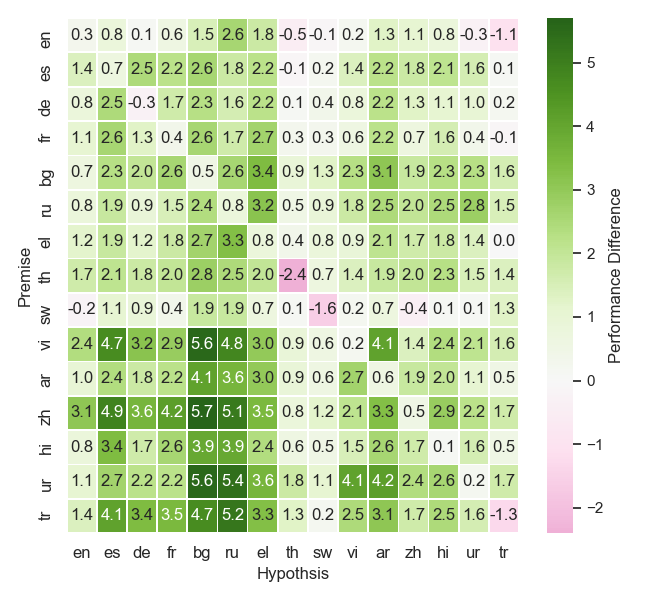}
    \caption{mBERT-RS-RP}
    \label{fig:heat-xnli-aug}
\end{subfigure}
\hfill
\begin{subfigure}[b]{0.49\textwidth}
    \centering
    \includegraphics[width=\textwidth]{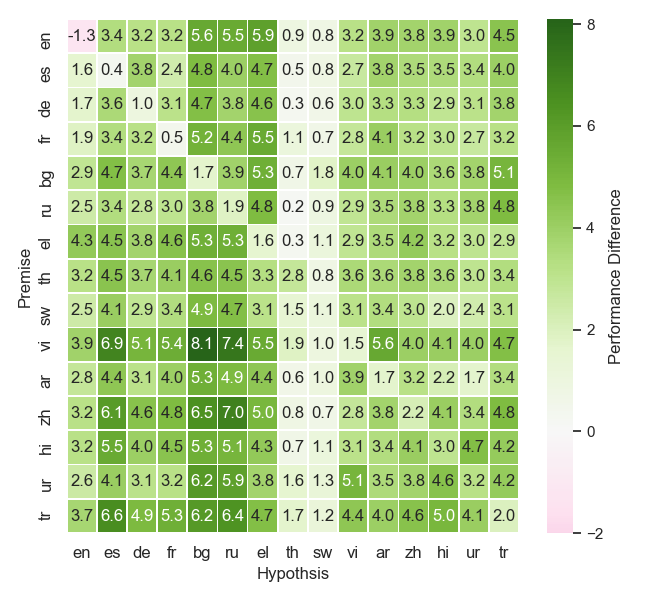}
    \caption{mBERT-RS-DA}
    \label{fig:heat-xnli-radv}
\end{subfigure}
\caption{Results for generalized zero-shot cross-lingual transfer on XNLI. We report the performance difference between the compared model and mBERT over different combinations of languages.}
\label{fig:heat-xnli}
\end{figure*}

\paragraph{How many augmented data needed for randomized smoothing?}

Since mBERT-RS-DA seems to be the most effective model for both PAWS-X and XNLI, we do further ablation on the number of augmented data for each example $m$. Figure~\ref{fig:aug} shows the average performance of mBERT-RS-DA on PAWS-X over different choices of $m$. We can observe that larger $m$ leads to better performance in general because more augmented examples help the model better approximate the local smoothness, resulting in more accurate robust regions. Interestingly, when $m\leq 10$, increasing $m$ can significantly improve the performance. When $m>10$, increasing $m$ only slightly improves the performance. This result suggests that setting $m$ to 10 for PAWS-X. Interestingly, we observe that setting $m$ to 3 is good enough for XNLI. This ablation study indicates that randomized smoothing with data augmentation can use just a few augmented instances per example to learn good robust regions.

\begin{table*}[t]
\centering
\setlength{\tabcolsep}{7pt}
\begin{tabular}{lcccccccc}
    \toprule
    Model       & en   & de   & es   & fr   & ja   & ko   & zh   & avg. \\
    \midrule
    mBERT*            & \textbf{94.0} & 85.7 & 87.4 & 87.0 & 73.0 & 69.6 & 77.0 & 82.0 \\
    mBERT (reproduce) & 93.7 & 85.4 & 88.2 & 87.8 & 75.3 & 74.2 & 79.1 & 83.4 \\
    mBERT-RS-DA       & 93.5 & \underline{\textbf{87.8}} & \textbf{88.8} & \underline{88.8} & \underline{\textbf{79.3}} & \underline{\textbf{78.3}} & \underline{81.5} & \underline{85.4} \\
    \midrule
    mBERT-RS-syntax   & 93.0 & 85.5 & 87.7 & 88.0 & \underline{76.5} & \underline{76.7} & \underline{80.7} & 83.5 \\
    \bottomrule
\end{tabular}
\caption{Results of syntactic perturbations on PAWS-X. Highest scores are in bold. Underlines denote that the improvement is significant with $p \leq 0.05$ for the bootstrapped paired $t$-test. *We report the numbers in the previous paper (\citet{Hu20xtreme}).}
\label{tab:syntax}
\end{table*}

\subsection{Zero-Shot Generalized Cross-Lingual Transfer Results}

Next, we study the zero-shot cross-lingual transfer in a \emph{generalized} setting.
\citet{Lewis20mlqa} proposed the generalized setting for the question answering task where the question and the context may belong to two different languages.\footnote{QA systems should be able to answer questions written in French by reading an English context.}
We consider the generalized setting for cross-lingual text classification since the input of PAWS-X and XNLI tasks are pairs of sentences.
For example, consider XNLI on English-Arabic sentence pairs; the premises are in English, and the hypotheses are in Arabic.
Note that due to the parallel nature of PAWS-X and XNLI dataset\footnote{PAWS-X and XNLI datasets consist of 7-way and 15-way parallel sentence pairs.}, we can pair up sentences from two different languages.
Notice that we directly use the trained models in Section~\ref{sec:results} to conduct inference in the generalized setting. In other words, all the classifiers are trained on English-English sentence pairs, without the consideration of target languages.

The results of mBERT-RS-RP and mBERT-RS-DA on PAWS-X and XNLI over all combinations of languages are shown in Figure~\ref{fig:heat-pawsx} and Figure~\ref{fig:heat-xnli}, respectively. 
While the diagonal numbers indicate the transfer results in the cross-lingual transfer settings, the non-diagonal entries present the generalized transfer performances.
Note that we report the performance difference between the compared model and mBERT (exact numbers can be found in Appendix~\ref{app:details}) and the languages are sorted according to their distances to English. 
We observe that the non-diagonal numbers are much larger than the diagonal numbers, which suggests that robust training results in larger performance improvements in the generalized cross-lingual transfer setting.
Given that the input sentences in training examples are in the same language (English), during inference, mBERT makes more mistakes in the classification tasks as the contextual representations for the input sentences may not be aligned accurately.
However, mBERT-RS-RP and mBERT-RS-DA can tolerate a certain amount of noise in input embeddings. Therefore, they are more stable when the input sentences come from different languages, leading to a significant improvement.

\subsection{Study on Syntactic Perturbations}
\label{sec:syntax}

As mentioned in Section~\ref{sec:method}, our primary focus is on the perturbations in the multilingual embedding space and does not consider the influence of language syntax in cross-lingual transfer. Different languages have linguistic differences, such as word order.
Differences in word order across languages affect the contextual embedding space that impacts cross-lingual transfer \cite{Ahmad19difficult}.
Therefore, we conduct a preliminary experiment to study the influence of syntax in robust training. 

mBERT-RS-DA uses a predefined synonym set to generate perturbed examples for data augmentation. Following a similar strategy, we construct \emph{syntactically perturbed examples} for data augmentation. More specifically, for every example $x = (w_1, w_2, ..., w_n)$ in $X_{src}$, we generate $m$ syntactically perturbed examples by randomly swapping adjacent words with a probability $p=0.1$. This random swapping may result in some examples with different word orders, which simulates the syntactic perturbations.
Then, we use those syntactically perturbed examples to train the smoothed classifier $g$, called mBERT-RS-syntax.

Table~\ref{tab:syntax} presents the preliminary results. The average performance of mBERT-RS-syntax is similar to the performance of standard mBERT. Interestingly, the zero-shot cross-lingual transfer performance drops when the target languages are more similar to the source language English (German, Spanish, and French), while the transfer performance increases when the target languages are more different from English (Japanese, Korean, and Chinese). This preliminary result suggests that it is possible to improve the zero-shot cross-lingual transfer by considering syntactic perturbations. One potential extension is adopting paraphrase generation models \cite{Iyyer18scpn,Huang21synpg} to construct more sophisticated syntactic perturbations and we leave this direction for future work.

\section{Conclusion}
In this work, we propose a robust model by drawing connections between adversarial examples and the failure cases of zero-shot cross-lingual transfer. We adopt two robust training methods, adversarial training and randomized smoothing, to train the desired robust model. The experimental results demonstrate that robust training improves zero-shot cross-lingual transfer on text classification tasks. In addition, the improvement is more significant in the generalized cross-lingual transfer setting.

\section*{Acknowledgments}
We thank anonymous reviewers for their helpful feedback. We thank UCLA-NLP group for the valuable discussions and comments. This work is supported in part by a Google Research Scholar Award, an Amazon Research Award, and the Intelligence Advanced Research Projects Activity (IARPA) via Contract No. 2019-19051600007.

\bibliography{emnlp2021}
\bibliographystyle{acl_natbib}

\clearpage

\appendix
\section{Detailed Results of Zero-Shot Generalized Cross-Lingual Transfer}
\label{app:details}

Table \ref{tab:mbert-pawsx}, \ref{tab:radv-pawsx}, and \ref{tab:aug-pawsx} show the result for mBERT, mBERT-RS-RP, mBERT-RS-DA on PAWS-X, respectively, while 
Table \ref{tab:mbert-xnli}, \ref{tab:radv-xnli}, and \ref{tab:aug-xnli} list the result for mBERT, mBERT-RS-RP, mBERT-RS-DA on XNLI, respectively. From those tables, we can observe that mBERT-RS-RP and mBERT-RS-DA lead to remarkable improvements.

\begin{table}[ht]
\centering
\small
\resizebox{\columnwidth}{!}{
\setlength{\tabcolsep}{3.5pt}
\begin{tabular}{c|ccccccc|c}
    \toprule
     & en & es & de & fr & zh & ko & ja & avg. \\
    \midrule
en & 93.7 & 85.4 & 85.0 & 85.0 & 66.5 & 66.4 & 63.4 & 77.9 \\
es & 86.1 & 88.2 & 80.5 & 83.9 & 63.7 & 64.0 & 60.9 & 75.3 \\
de & 85.6 & 79.7 & 85.4 & 79.8 & 63.9 & 64.7 & 61.5 & 74.4 \\
fr & 84.8 & 83.0 & 80.3 & 87.8 & 63.7 & 63.9 & 61.1 & 74.9 \\
zh & 66.7 & 63.7 & 63.9 & 64.3 & 79.1 & 62.4 & 64.3 & 66.3 \\
ko & 67.1 & 64.9 & 65.3 & 65.0 & 62.7 & 74.2 & 65.1 & 66.3 \\
ja & 63.0 & 61.1 & 61.1 & 61.1 & 64.6 & 63.6 & 75.3 & 64.3 \\
    \midrule
avg. & 78.1 & 75.1 & 74.5 & 75.3 & 66.3 & 65.6 & 64.5 & 71.3 \\
    \bottomrule
\end{tabular}}
\caption{Results for mBERT on PAWS-X.}
\label{tab:mbert-pawsx}
\end{table}

\begin{table}[ht]
\centering
\small
\resizebox{\columnwidth}{!}{
\setlength{\tabcolsep}{3.5pt}
\begin{tabular}{c|ccccccc|c}
    \toprule
     & en & es & de & fr & zh & ko & ja & avg. \\
    \midrule
en & 94.5 & 89.3 & 88.2 & 88.9 & 74.9 & 70.8 & 70.6 & 82.5 \\
es & 89.3 & 90.0 & 84.6 & 87.7 & 72.5 & 68.6 & 67.5 & 80.0 \\
de & 88.7 & 84.2 & 87.4 & 84.4 & 72.0 & 69.5 & 68.8 & 79.3 \\
fr & 88.6 & 86.9 & 83.9 & 89.5 & 71.8 & 69.3 & 68.3 & 79.7 \\
zh & 74.6 & 72.3 & 71.9 & 72.4 & 82.0 & 69.9 & 72.0 & 73.6 \\
ko & 70.1 & 68.4 & 68.0 & 69.0 & 69.6 & 77.5 & 72.5 & 70.7 \\
ja & 69.7 & 67.7 & 67.5 & 67.9 & 72.5 & 72.1 & 77.9 & 70.7 \\
    \midrule
avg. & 82.2 & 79.8 & 78.8 & 80.0 & 73.6 & 71.1 & 71.1 & 76.7 \\
    \bottomrule
\end{tabular}}
\caption{Results for mBERT-RS-RP on PAWS-X.}
\label{tab:radv-pawsx}
\end{table}

\begin{table}[ht]
\centering
\small
\resizebox{\columnwidth}{!}{
\setlength{\tabcolsep}{3.5pt}
\begin{tabular}{c|ccccccc|c}
    \toprule
     & en & es & de & fr & zh & ko & ja & avg. \\
    \midrule
en & 93.5 & 89.1 & 89.1 & 89.1 & 75.6 & 72.2 & 72.3 & 83.0 \\
es & 89.7 & 88.8 & 85.7 & 87.5 & 73.0 & 70.0 & 70.1 & 80.7 \\
de & 89.5 & 85.1 & 87.9 & 85.5 & 72.8 & 70.7 & 70.7 & 80.3 \\
fr & 89.2 & 86.9 & 85.5 & 88.8 & 73.4 & 70.5 & 70.3 & 80.7 \\
zh & 76.1 & 73.3 & 73.6 & 73.9 & 81.5 & 71.8 & 74.2 & 74.9 \\
ko & 72.0 & 70.4 & 70.3 & 70.5 & 71.5 & 78.3 & 74.1 & 72.4 \\
ja & 71.3 & 69.2 & 70.2 & 69.8 & 72.8 & 73.2 & 79.3 & 72.3 \\
    \midrule
avg. & 83.0 & 80.4 & 80.3 & 80.7 & 74.4 & 72.4 & 73.0 & 77.7 \\
    \bottomrule
\end{tabular}}
\caption{Results for mBERT-RS-DA on PAWS-X.}
\label{tab:aug-pawsx}
\end{table}

\begin{table*}[t]
\centering
\small
\aboverulesep = 0.3mm
\belowrulesep = 0.3mm
\resizebox{0.97\textwidth}{!}{
\setlength{\tabcolsep}{3.5pt}
\begin{tabular}{c|ccccccccccccccc|c}
    \toprule
    & en & es & de & fr & bg & ru & el & th & sw & vi & ar & zh & hi & ur & tr & avg. \\
    \midrule
en & 82.3 & 70.3 & 65.8 & 69.7 & 60.5 & 63.1 & 55.3 & 44.6 & 41.1 & 63.9 & 57.7 & 64.6 & 52.0 & 49.5 & 52.3 & 59.5 \\
es & 73.5 & 74.3 & 62.9 & 69.0 & 60.5 & 63.7 & 57.3 & 44.6 & 40.6 & 61.4 & 57.9 & 60.8 & 50.4 & 47.1 & 51.6 & 58.4 \\
de & 71.8 & 65.5 & 70.8 & 65.6 & 59.5 & 63.3 & 55.8 & 44.3 & 41.0 & 60.2 & 56.5 & 60.1 & 52.5 & 49.4 & 52.0 & 57.9 \\
fr & 73.6 & 69.0 & 64.0 & 73.8 & 59.5 & 63.1 & 55.7 & 44.1 & 40.5 & 62.2 & 57.3 & 61.6 & 51.1 & 48.5 & 51.8 & 58.4 \\
bg & 67.8 & 63.7 & 60.8 & 62.5 & 68.2 & 64.2 & 56.0 & 44.2 & 39.9 & 57.4 & 56.3 & 57.8 & 51.2 & 47.2 & 50.3 & 56.5 \\
ru & 69.1 & 65.2 & 62.6 & 64.4 & 62.7 & 68.7 & 55.0 & 44.2 & 39.9 & 59.0 & 56.7 & 58.6 & 50.6 & 46.8 & 50.0 & 56.9 \\
el & 62.7 & 61.4 & 58.0 & 60.2 & 57.1 & 57.7 & 66.4 & 44.4 & 40.5 & 56.4 & 55.6 & 54.0 & 49.6 & 46.8 & 50.7 & 54.8 \\
th & 54.8 & 52.0 & 49.9 & 51.3 & 49.1 & 50.4 & 49.0 & 53.0 & 39.4 & 51.1 & 49.9 & 49.3 & 45.9 & 44.8 & 45.4 & 49.0 \\
sw & 54.2 & 51.2 & 48.7 & 50.5 & 47.2 & 47.9 & 47.9 & 41.8 & 50.0 & 48.5 & 49.1 & 48.5 & 45.4 & 44.4 & 45.8 & 48.1 \\
vi & 67.4 & 60.3 & 57.4 & 61.2 & 52.9 & 57.1 & 52.9 & 44.2 & 39.8 & 70.3 & 53.3 & 62.0 & 49.2 & 45.9 & 47.5 & 54.8 \\
ar & 63.9 & 60.4 & 57.0 & 59.5 & 54.5 & 57.1 & 53.3 & 43.9 & 40.4 & 55.4 & 64.8 & 55.2 & 50.3 & 48.4 & 49.9 & 54.3 \\
zh & 67.9 & 59.9 & 57.2 & 59.9 & 53.4 & 56.5 & 50.4 & 42.7 & 39.6 & 60.8 & 53.5 & 69.2 & 48.0 & 45.7 & 48.0 & 54.2 \\
hi & 61.4 & 55.5 & 55.0 & 55.3 & 52.6 & 54.4 & 51.9 & 43.8 & 40.3 & 53.8 & 53.1 & 53.7 & 59.7 & 52.7 & 49.9 & 52.9 \\
ur & 60.1 & 54.0 & 53.9 & 55.1 & 48.8 & 51.5 & 49.6 & 41.9 & 39.7 & 50.0 & 52.1 & 52.3 & 54.4 & 57.7 & 48.2 & 51.3 \\
tr & 61.0 & 55.1 & 53.6 & 55.1 & 52.0 & 52.6 & 50.9 & 42.4 & 40.7 & 52.3 & 52.0 & 53.2 & 49.7 & 47.3 & 60.9 & 51.9 \\
    \midrule
avg. & 66.1 & 61.2 & 58.5 & 60.9 & 55.9 & 58.1 & 53.8 & 44.3 & 40.9 & 57.5 & 55.1 & 57.4 & 50.7 & 48.1 & 50.3 & 54.6 \\
    \bottomrule
\end{tabular}}
\caption{Results for mBERT on XNLI.}
\label{tab:mbert-xnli}
\end{table*}

\begin{table*}[t]
\centering
\small
\aboverulesep = 0.3mm
\belowrulesep = 0.3mm
\resizebox{0.97\textwidth}{!}{
\setlength{\tabcolsep}{3.5pt}
\begin{tabular}{c|ccccccccccccccc|c}
    \toprule
    & en & es & de & fr & bg & ru & el & th & sw & vi & ar & zh & hi & ur & tr & avg. \\
    \midrule
en & 82.6 & 71.2 & 65.9 & 70.3 & 62.0 & 65.7 & 57.0 & 44.1 & 40.9 & 64.1 & 58.9 & 65.7 & 52.8 & 49.2 & 51.2 & 60.1 \\
es & 74.9 & 75.0 & 65.4 & 71.2 & 63.0 & 65.6 & 59.5 & 44.5 & 40.8 & 62.9 & 60.1 & 62.6 & 52.5 & 48.7 & 51.7 & 59.9 \\
de & 72.6 & 68.0 & 70.5 & 67.4 & 61.7 & 64.9 & 58.0 & 44.4 & 41.4 & 61.0 & 58.8 & 61.4 & 53.6 & 50.4 & 52.2 & 59.1 \\
fr & 74.7 & 71.6 & 65.2 & 74.1 & 62.1 & 64.8 & 58.4 & 44.4 & 40.8 & 62.7 & 59.5 & 62.4 & 52.7 & 48.9 & 51.7 & 59.6 \\
bg & 68.5 & 66.0 & 62.9 & 65.1 & 68.7 & 66.8 & 59.4 & 45.1 & 41.1 & 59.7 & 59.4 & 59.7 & 53.5 & 49.6 & 51.8 & 58.5 \\
ru & 69.9 & 67.1 & 63.5 & 65.9 & 65.0 & 69.5 & 58.2 & 44.7 & 40.9 & 60.8 & 59.2 & 60.6 & 53.1 & 49.5 & 51.5 & 58.6 \\
el & 63.9 & 63.3 & 59.3 & 62.0 & 59.8 & 61.0 & 67.2 & 44.7 & 41.3 & 57.3 & 57.7 & 55.7 & 51.4 & 48.2 & 50.7 & 56.2 \\
th & 56.4 & 54.1 & 51.7 & 53.3 & 51.9 & 52.9 & 51.0 & 50.5 & 40.1 & 52.5 & 51.8 & 51.3 & 48.2 & 46.3 & 46.8 & 50.6 \\
sw & 54.1 & 52.3 & 49.6 & 50.9 & 49.1 & 49.7 & 48.6 & 41.8 & 48.4 & 48.7 & 49.8 & 48.1 & 45.4 & 44.5 & 47.1 & 48.6 \\
vi & 69.9 & 65.0 & 60.5 & 64.1 & 58.6 & 61.8 & 56.0 & 45.1 & 40.4 & 70.5 & 57.4 & 63.4 & 51.5 & 48.0 & 49.1 & 57.4 \\
ar & 64.9 & 62.8 & 58.7 & 61.7 & 58.7 & 60.7 & 56.3 & 44.7 & 41.1 & 58.1 & 65.4 & 57.1 & 52.2 & 49.6 & 50.3 & 56.1 \\
zh & 71.1 & 64.8 & 60.8 & 64.1 & 59.0 & 61.6 & 53.9 & 43.5 & 40.8 & 63.0 & 56.8 & 69.7 & 50.9 & 47.8 & 49.7 & 57.2 \\
hi & 62.2 & 58.9 & 56.7 & 57.9 & 56.5 & 58.3 & 54.3 & 44.5 & 40.8 & 55.3 & 55.8 & 55.3 & 59.8 & 54.2 & 50.4 & 54.7 \\
ur & 61.2 & 56.7 & 56.1 & 57.3 & 54.4 & 57.0 & 53.2 & 43.7 & 40.8 & 54.1 & 56.3 & 54.6 & 56.9 & 57.9 & 49.9 & 54.0 \\
tr & 62.4 & 59.2 & 57.0 & 58.6 & 56.7 & 57.9 & 54.2 & 43.7 & 40.9 & 54.8 & 55.1 & 54.9 & 52.2 & 48.8 & 59.7 & 54.4 \\
    \midrule
avg. & 67.3 & 63.7 & 60.3 & 62.9 & 59.1 & 61.2 & 56.4 & 44.6 & 41.4 & 59.0 & 57.5 & 58.8 & 52.4 & 49.4 & 50.9 & 56.3 \\
    \bottomrule
\end{tabular}}
\caption{Results for mBERT-RS-RP on XNLI.}
\label{tab:radv-xnli}
\end{table*}

\begin{table*}[t]
\centering
\small
\aboverulesep = 0.3mm
\belowrulesep = 0.3mm
\resizebox{0.97\textwidth}{!}{
\setlength{\tabcolsep}{3.5pt}
\begin{tabular}{c|ccccccccccccccc|c}
    \toprule
    & en & es & de & fr & bg & ru & el & th & sw & vi & ar & zh & hi & ur & tr & avg. \\
    \midrule
en & 81.0 & 73.7 & 69.0 & 72.8 & 66.2 & 68.6 & 61.2 & 45.5 & 41.9 & 67.1 & 61.6 & 68.4 & 55.9 & 52.4 & 56.8 & 62.8 \\
es & 75.2 & 74.7 & 66.7 & 71.4 & 65.3 & 67.7 & 62.0 & 45.1 & 41.4 & 64.1 & 61.7 & 64.3 & 53.9 & 50.5 & 55.5 & 61.3 \\
de & 73.4 & 69.2 & 71.9 & 68.7 & 64.1 & 67.1 & 60.4 & 44.6 & 41.6 & 63.2 & 59.9 & 63.4 & 55.4 & 52.4 & 55.8 & 60.7 \\
fr & 75.4 & 72.4 & 67.2 & 74.2 & 64.7 & 67.5 & 61.2 & 45.2 & 41.2 & 65.0 & 61.4 & 64.8 & 54.2 & 51.2 & 55.0 & 61.4 \\
bg & 70.7 & 68.4 & 64.6 & 66.9 & 69.9 & 68.0 & 61.3 & 44.9 & 41.6 & 61.4 & 60.4 & 61.7 & 54.8 & 51.0 & 55.4 & 60.1 \\
ru & 71.6 & 68.6 & 65.4 & 67.4 & 66.4 & 70.6 & 59.7 & 44.4 & 40.9 & 61.9 & 60.2 & 62.4 & 53.9 & 50.5 & 54.8 & 59.9 \\
el & 67.0 & 65.9 & 61.9 & 64.8 & 62.3 & 63.0 & 68.0 & 44.6 & 41.6 & 59.3 & 59.1 & 58.2 & 52.7 & 49.7 & 53.7 & 58.1 \\
th & 58.0 & 56.4 & 53.6 & 55.4 & 53.7 & 54.9 & 52.4 & 55.8 & 40.3 & 54.7 & 53.5 & 53.0 & 49.5 & 47.8 & 48.8 & 52.5 \\
sw & 56.7 & 55.3 & 51.7 & 53.9 & 52.2 & 52.6 & 51.0 & 43.2 & 51.1 & 51.6 & 52.5 & 51.5 & 47.4 & 46.8 & 48.9 & 51.1 \\
vi & 71.3 & 67.2 & 62.4 & 66.5 & 61.0 & 64.5 & 58.5 & 46.1 & 40.8 & 71.8 & 58.9 & 66.0 & 53.3 & 49.9 & 52.2 & 59.4 \\
ar & 66.7 & 64.9 & 60.0 & 63.5 & 59.8 & 61.9 & 57.7 & 44.4 & 41.5 & 59.3 & 66.4 & 58.4 & 52.5 & 50.1 & 53.3 & 57.4 \\
zh & 71.2 & 66.0 & 61.9 & 64.7 & 59.9 & 63.5 & 55.4 & 43.5 & 40.3 & 63.6 & 57.3 & 71.5 & 52.1 & 49.1 & 52.8 & 58.2 \\
hi & 64.6 & 60.9 & 59.0 & 59.9 & 57.9 & 59.5 & 56.2 & 44.5 & 41.4 & 56.9 & 56.6 & 57.8 & 62.8 & 57.4 & 54.1 & 56.6 \\
ur & 62.8 & 58.1 & 57.0 & 58.3 & 55.0 & 57.5 & 53.5 & 43.6 & 41.0 & 55.1 & 55.6 & 56.1 & 58.9 & 60.9 & 52.4 & 55.0 \\
tr & 64.7 & 61.7 & 58.5 & 60.4 & 58.2 & 59.0 & 55.7 & 44.1 & 41.9 & 56.6 & 56.0 & 57.7 & 54.7 & 51.4 & 62.9 & 56.2 \\
    \midrule
avg. & 68.7 & 65.6 & 62.0 & 64.6 & 61.1 & 63.1 & 58.3 & 45.3 & 41.9 & 60.8 & 58.7 & 61.0 & 54.1 & 51.4 & 54.2 & 58.0 \\
    \bottomrule
\end{tabular}}
\caption{Results for mBERT-RS-DA on XNLI.}
\label{tab:aug-xnli}
\end{table*}

\end{document}